\documentclass[graybox]{svmult}

\usepackage{mathptmx}       
\usepackage{helvet}         
\usepackage{courier}        
\usepackage{type1cm}        
\usepackage{makeidx}         
\usepackage{graphicx}        
\usepackage{booktabs}
\usepackage{multicol}        
\usepackage[bottom]{footmisc}
\usepackage{hyperref}

\makeindex             

\begin{document}

\title*{Multilingual Question Answering applied to Conversational Agents}
\author{Wissam Siblini, Charlotte Pasqual, Axel Lavielle, Mohamed Challal and Cyril Cauchois}
\authorrunning{Wissam Siblini and al.}
\institute{Wissam Siblini, Charlotte Pasqual, Axel Lavielle, Mohamed Challal and Cyril Cauchois \at Worldline, Lyon, France \email{firstname.lastname@worldline.com}}
%
%
\maketitle

\abstract*{Recent advances with language models (e.g. BERT, XLNet, ...), have allowed surpassing human performance on complex NLP tasks such as Reading Comprehension. However, labeled datasets for training are available mostly in English which makes it difficult to acknowledge progress in other languages. Fortunately, models are now pre-trained on unlabeled data from hundreds of languages and exhibit interesting transfer abilities from one language to another. In this paper, we show that multilingual BERT is naturally capable of zero-shot transfer for an extractive Question Answering task (eQA) from English to other languages. More specifically, it outperforms the best previously known baseline for transfer to Japanese and French. Moreover, using a recently published large eQA French dataset, we are able to further show that (1) zero-shot transfer provides results really close to a direct training on the target language and (2) combination of transfer and training on target is the best option overall. We finally present a practical application: a multilingual conversational agent called Kate which answers to HR-related questions in several languages directly from the content of intranet pages.}

\abstract{Recent advances with language models (e.g. BERT, XLNet, ...), have allowed surpassing human performance on complex NLP tasks such as Reading Comprehension. However, labeled datasets for training are available mostly in English which makes it difficult to acknowledge progress in other languages. Fortunately, models are now pre-trained on unlabeled data from hundreds of languages and exhibit interesting transfer abilities from one language to another. In this paper, we show that multilingual BERT is naturally capable of zero-shot transfer for an extractive Question Answering task (eQA) from English to other languages. More specifically, it outperforms the best previously known baseline for transfer to Japanese and French. Moreover, using a recently published large eQA French dataset, we are able to further show that (1) zero-shot transfer provides results really close to a direct training on the target language and (2) combination of transfer and training on target is the best option overall. We finally present a practical application: a multilingual conversational agent called Kate which answers to HR-related questions in several languages directly from the content of intranet pages.}

\section{Introduction}
\label{p4_sec:introduction}
Over the past few years, we have witnessed a revolution of machine learning in the domain of Natural Language Processing (NLP) \cite{vaswani2017attention,devlin2018bert}. Perhaps motivated by needs and competitions (e.g. GLUE \cite{wang2018glue}), novel and powerful proposals are made every day by public and private laboratories around the world to solve several complex NLP tasks: Natural Language Inference \cite{williams2017broad,levesque2012winograd}, Sentence Similarity and Paraphrasing \cite{dolan2005automatically,agirre2012semeval}, Text Classification \cite{socher2013recursive,warstadt2018neural}, Reading Comprehension/Question Answering (QA) \cite{rajpurkar2016squad,nguyen2016ms,lai2017race,joshi2017triviaqa}. For instance, trained on the SQuAD dataset \cite{rajpurkar2016squad}, language models such as BERT \cite{devlin2018bert}, RoBERTa \cite{liu2019roberta} or XLNet \cite{yang2019xlnet}, have shown a great ability in identifying the answer to a question in a given unstructured source of information (raw text). For most conversational agents currently limited to the detection of a user's intent among a predefined set of possibility, this capability offers new interesting features. In particular, developers would no longer need to anticipate everything as the agent would be able to answer to unexpected questions from web pages or a documentation. With today's available pretrained models and datasets, this feature can be implemented with success for English but, even for other widely used languages, data/models remain scarce. Could we still make it work for those languages then? Creating labeled QA datasets in every target languages would be resource consuming and not very flexible. An alternative direction is transfer learning and in particular, zero-shot transfer \cite{hardalov2019beyond,liu2019xqa} when there is no target data. On the one hand, many specific strategies have been proposed for zero-shot transfer with explicit language alignment \cite{firat2016zero,johnson2017google,asai2018multilingual}. On the other hand, state-of-the-art language models are currently being pre-trained and released on hundreds of languages and they seem to naturally integrate language alignment which allows a surprisingly good performance in zero-shot transfer. 

In this paper, after introducing the history behind \textit{Transformer}-based language models and then the specificity of BERT and its multilingual version, we empirically demonstrate the latter's capability, when trained to solve the SQuAD task in English, to do the same in other languages (French and Japanese) without using any labeled target data. We also introduce six new cross-lingual QA datasets (question and source in a different language) to better understand the mechanism behind the transfer ability. We then show that we can further improve the results using "virtual" target data obtained with machine translation and end up with a performance competitive to the one that we would have with target data annotated by humans \cite{fquad}. We finally present a practical use of the trained models: an HR virtual assistant named Kate able to automatically answer to user questions from multilingual web pages.

\section{Related Works: Language models for Natural Language Processing}
\label{p4_sec:related}

Natural Language Processing (NLP) is a key application of machine learning. Text Classification was one of its first tasks where results were very promising (e.g. for spam detection \cite{pantel1998spamcop}) with strategies combining heuristics for pre-processing (e.g. TF-IDF \cite{jones2004statistical}) and multi-class classifiers such as SVM \cite{manevitz2001one}. Originally, proposals restricted the model's training to the dataset specifically gathered for the targeted task. Later, researchers proposed to pre-learn a language representation on large external unsupervised text sources to provide models with semantic information. The most emblematic example is word2vec \cite{mikolov2013distributed}, a simple neural network (one hidden layer), trained with several tricks to predict a word in a sentence from its context (surrounding words) and vice versa. Once the network is trained, word representations/embeddings are derived from the network's weights. Because they reflect the semantic of the language (e.g. vectors of synonyms have a high similarity), they have successfully been employed to improve the ability of models to generalize on target tasks even with limited sized labeled datasets \cite{joulin2016bag}. Therefore, they have been largely adopted and then incrementally evolved to more complex sentence/paragraph embeddings \cite{kiros2015skip,logeswaran2018efficient,le2014distributed}, with recurrent neural networks for instance. Researches also focused on building context-dependent embeddings for disambiguation \cite{peters2018deep}. 

Over the last two years, proposals have been geared towards standardizing all these representation/transfer learning progresses into a format easily usable by practitioners for solving NLP tasks: the so-called language models. In these deep models, text representation is integrated within the first layers (representation layers), and the output layers are changeable to offer the possibility of solving a wide variety of tasks \cite{devlin2018bert, howard2018universal}. The input is the text represented as a sequence of "tokens" (characters or series of characters) \cite{schuster2012japanese}. The representation layers are pre-trained on unlabeled text data to learn generic and contextualized representations of each elements of the input sequence. Then, given a target task, the output layers are customized and the whole model is fine-tuned on target data. Recent language models such as BERT rely on this principle and are generally inspired from the architecture of the \textit{Transformer} \cite{vaswani2017attention}, a neural network with several layers of attention and self-attention, able to solve difficult NLP tasks by modeling complex interdependencies between items of long sequences of text.

\section{BERT and its multilingual extension}

\begin{figure}[h]
\centering
\includegraphics[width=0.9\textwidth]{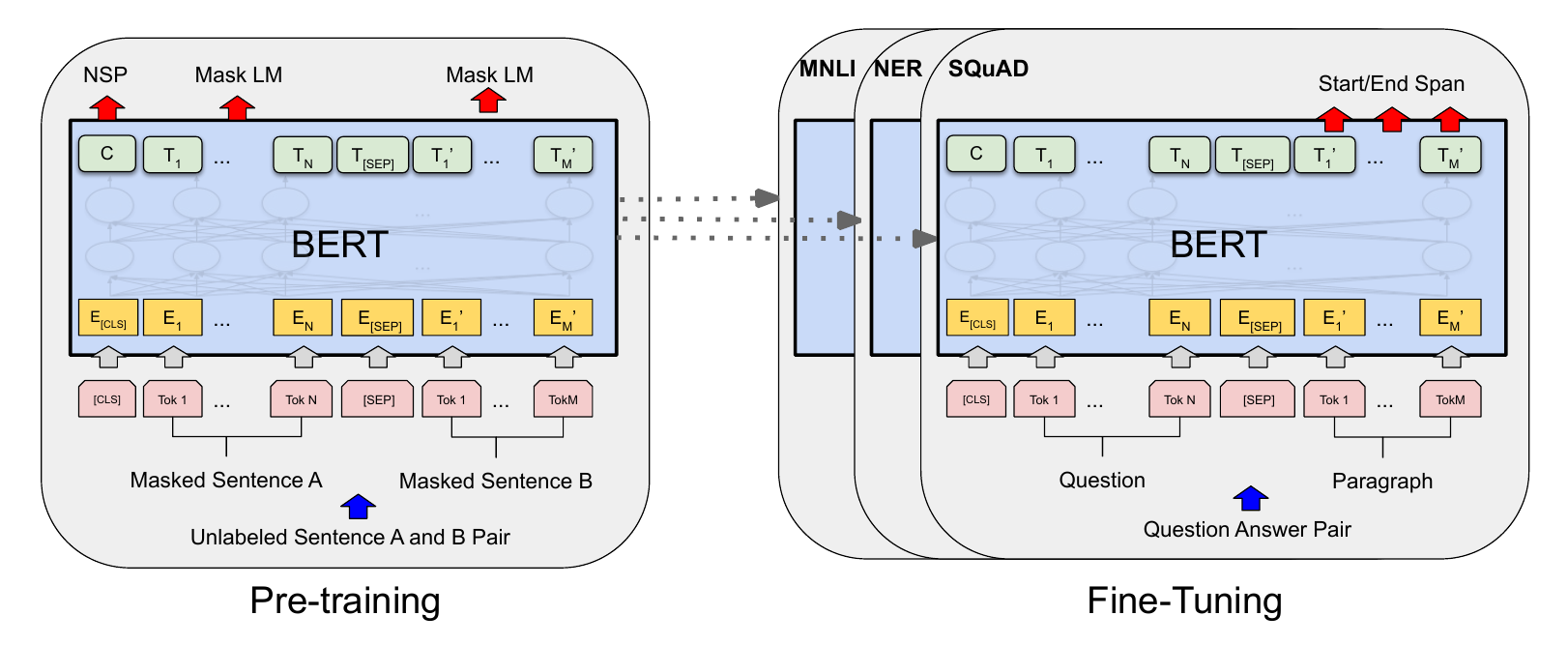}
\caption{BERT's global structure - Source: \cite{devlin2018bert}}\label{p4_figure_bert}
\end{figure}

Several language models have a competitive performance on the question answering task that we target in this paper. In the following, we choose to focus on BERT as it was published in a largely multilingual version. BERT learns representations of tokens jointly contextualized by their left context and their right context using the \textit{Transformer}'s encoder. It is designed to take as input either a single sentence or a pair of sentences separated by a special token (figure \ref{p4_figure_bert}). It has a wide range of outputs allowing it to be adapted for several tasks (classification, text generation, question-answering) without significant modification of its architecture.

Its pre-training consists in the learning of two tasks. In the first, about $15\%$ of the tokens, sampled randomly, are masked in the input (by replacing them with the special token [MASK], a random token or the unchanged token) and the objective is to retrieve them. In the second, the goal is to predict, given two sentences A and B, if sentence B directly follows sentence A in the corpus or not. For the multilingual version, the pre-training corpus is composed of unlabeled documents from Wikipedia, in the hundred languages with the largest number of articles\footnote{\url{https://meta.wikimedia.org/wiki/List_of_Wikipedias}}. A sampling of the articles and a weighting of the tokens, based on the frequency of the language, was performed to balance languages to some extent. The model is intentionally not informed of the language of the sample, so the representations of the tokens learned are not explicitly specific to a language.  

BERT pre-training is very expensive: four days on four to sixteen cloud TPUs. Fortunately, the authors shared the pre-trained weights on their github \footnote{\url{https://github.com/google-research/bert}}. On the contrary, the fine-tuning on specific tasks is relatively quick. For example, on SQuAD (more than one hundred thousand samples), it takes about two hours on a conventional GPU (Tesla V100). BERT has been published in several versions: the main ones are base and large. The difference lies in the size of the model (size of the hidden layers, number of blocks and number of self-attention heads). The large version is slightly more accurate on NLP tasks but only the base version can be trained on SQuAD with a standard GPU (12 GB of RAM). 

\section{Experiments on multilingual and cross-lingual QA tasks}

In this section, we study the ability of multilingual bert (mBERT) for zero-shot transfer on an extractive question answering task. We base our empirical analysis on the SQuAD question-answering format: given a question-paragraph pair, the goal is to determine the location of the answer in the paragraph. We assess if mBERT, once trained on English SQuAD, is able to solve the same task in French, Japanese, and on cross-lingual data. Then, using a recently published dataset in French called FQuAD, we compare this zero shot transfer performance to the one that mBERT would obtain if it was natively trained in French.

\subsection{Experiment 1: zero-shot transfer on French and Japanese}

A sample of the English SQuAD v1.1 test set (only the first paragraph of each of the 48 Wikipedia pages) has been translated by humans in French and Japanese and made available online\footnote{\url{https://github.com/AkariAsai/extractive_rc_by_runtime_mt}}. 

\begin{figure}[h]
\centering
\includegraphics[width=0.8\textwidth]{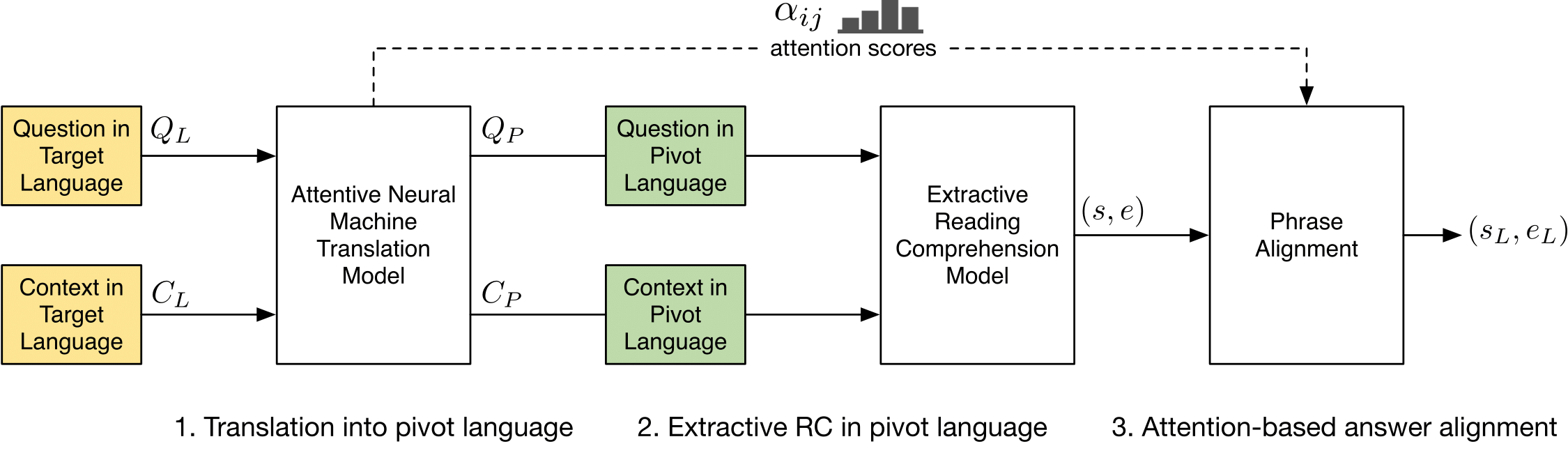}
\caption{Illustration of the architecture of the baseline in three parts - Source: \cite{asai2018multilingual}}\label{p4_fig_asai_method}
\end{figure}

We start by fine-tuning mBERT on the English SQuAD train set (we use the cased multilingual base model and the training hyperparameters specified in the official github repository of BERT). Then we evaluate its performance on the French and Japanese sets and compare them to the baseline \cite{asai2018multilingual} with the best results published to date. The baseline explicitly combines two models: one language-specific model for Reading Comprehension (RC) in a pivot language (English) and an attentive Machine-Translation model (MT) between the pivot language and the target language (Japanese or French). The algorithm is summarized in figure \ref{p4_fig_asai_method}. First, the MT model translates the passage-question pair from the target language to the pivot language, then the RC model extracts the answer in the pivot language and, the algorithm finally recovers the answer in the original language using soft-alignment attention scores from the MT model.

Table \ref{p4_results_vs_baseline} displays the Exact Match (EM) and F1-score of multilingual BERT, which significantly surpass those of the baseline for Japanese and French. In addition, note that mBERT has the structural additional advantage of being more easily transferable to even other languages. 
\begin{table}[ht]
\centering
\caption{\label{p4_results_vs_baseline} Comparison of Exact Match and F1-score of mBERT and the baseline on French and Japanese SQuAD. F1 and EM are the two official metrics of the SQuAD benchmark. EM measures the percentage of predictions that match exactly the ground-truth location of the answer. F1 measures the average overlap between the prediction and ground truth answer.}
\vspace{0.2cm}
\begin{tabular}[t]{ccccc}
\toprule
 &\multicolumn{2}{c}{French}&\multicolumn{2}{c}{Japanese}\\
&F1&EM&F1&EM\\
\midrule
Baseline&61.88&40.67&52.19&37.00\\
Multilingual BERT&\textbf{76.65}&\textbf{61.77}&\textbf{61.83}&\textbf{59.94}\\
\bottomrule
\end{tabular}
\end{table}

\subsection{Experiment 2: Cross lingual question answering}

To run cross-lingual tests, we build six additional datasets from the existing ones by mixing context in one language with question in another language\footnote{\url{https://github.com/wissam-sib/multilingualQA}}. The performance of mBERT on all datasets is displayed in Table \ref{p4_tab_cross_lingual}. Since the model was trained for the task in English, the performance is the best for the En-En dataset. The performance on Fr-Fr and Jap-Jap is also very good as noted in the first experiment. Results on cross-lingual sets are often close to monolingual results and in some cases, like on the En-Fr dataset, the performance is even higher (see comparison with the Fr-Fr dataset). We also point out that the exact match and F1-score are close together when the context is in Japanese whereas there is generally a larger gap for the other two languages. It could be due to the fact that Japanese tokens are bigger parts of words so there is less room for partial overlapping between predictions and ground-truth.

\begin{table}[ht]
\centering
   \caption{\label{p4_tab_cross_lingual} Exact Match and F1-score of mBERT on each of the cross-lingual SQuAD datasets. The row language is the one of the paragraph and the column language is the one of the question. The figures in bold are the best exact match, for each language, among the datasets where they occur.}
   \vspace{0.2cm}
   \begin{tabular}{cccccccc}
      \toprule
&Question&\multicolumn{2}{c}{En}&\multicolumn{2}{c}{Fr}&\multicolumn{2}{c}{Jap}\\ 
&&F1&EM&F1&EM&F1&EM\\ 
\midrule
&En&90.57&\textbf{81.96}&78.55&\textbf{67.28}&66.22&52.91\\ 
Context&Fr&81.10&65.14&76.65&61.77&60.28&42.20\\ 
&Jap&58.95&57.49&47.19&45.26&61.83&\textbf{59.93}\\ 
\bottomrule
   \end{tabular}
\end{table}

\subsection{Experiment 3: Comparison between the transferred model and a target model}

In early 2020, FQuAD 1.0, the first large scale set of French native questions and answers samples annotated by humans (about 25k samples) was released \cite{fquad}. Here, we use it to analyse whether zero shot transfer from English is competitive, in a target language (French), with a fully supervised learning on target data. More specifically, we compare the performance in French of the same model (mBERT), with the same hyperparameters, but either trained on English SQuAD (zero-shot transfer) or on the French FQuAD 1.0 train set (target baseline). The evaluation is done on both the FQuAD 1.0 dev set and on the French sample of SQuAD 1.1 dev set (introduced in Experiment 1). To further improve the zero-shot approach, we also consider a training on both the English SQuAD and a "French" SQuAD (Fr-SQuAD 1.1) obtained by automatic translation of SQuAD 1.1 using Google's API\footnote{\url{https://github.com/Alikabbadj/French-SQuAD}}.

\begin{table}[ht]
\centering
   \caption{\label{p4_tab_trainings} Exact Match and F1-score of mBERT (trained on different training sets) on the FQuAD 1.0 dev set and the French SQuAD 1.1 test set. Each row is a different training set. The figures in bold are the best scores, for each test set, among all the trainings.}
   \vspace{0.2cm}
   \begin{tabular}{ccccc}
      \toprule
&\multicolumn{2}{c}{FQuAD}&\multicolumn{2}{c}{French Test Set}\\ 
Training Set&F1&EM&F1&EM\\ 
\midrule
SQuAD 2.0 (Zero-shot transfer)&71.43&47.65&76.65&61.77\\ 
FQuAD 1.0 (Target baseline) &74.75&49.87&77.11&61.16\\ 
SQuAD 2.0 + Fr-SQuAD 1.1 &73.90&46.42&82.39&\textbf{66.36}\\
SQuAD + Fr-SQuAD + FQuAD &\textbf{76.71}&\textbf{51.63}&\textbf{83.41}&\textbf{66.36}\\ 
\bottomrule
   \end{tabular}
\end{table}

We notice that the zero shot transfer approach (training on SQuAD 2.0 only) is almost as good as training on FQuAD. Additionally using SQuAD's automatic translation in French makes the model even better. Finally, using all datasets seems to be the best approach overall.

\subsection{Discussion}

Multilingual BERT, only fine-tuned on English data, seems very powerful for solving a QA task in three target languages. Our last experiment even shows that its zero shot transfer performance is almost as good as if it was trained directly in the target language, and can be further improved with automatic translation. This opens up interesting opportunities for languages suffering from a lack of task-oriented datasets. And even for target languages with available datasets, using both data in English and in the target language achieves better results than using only the target data.

The impressive zero-shot transfer capability of mBERT after a simple fine-tuning on the English SQuAD could be explained by the learning of four concepts overall: (1) the semantic structure of English and other languages, (2) the alignment between the embeddings of the different languages (at least for the most frequent ones \cite{Pires2019HowMI}), (3) the particularities of the question-answering task in English and (4) the intrinsic question-answering concepts that apply to all languages.

Assumptions (1) and (3) have already been widely validated in the past on the English SQuAD benchmark. Let us then focus on hypotheses (2) and (4) which are motivated by two intuitions. On the one hand, many common words, for instance named entities, are present in the Wikipedia articles of most languages. These could have played the role of landmarks in the pre-training phase and allowed the model to align the spaces of representation of the different languages. On the other hand, certain characteristics of the question-answering task are certainly independent of the language, such as the identification of passages in the paragraph where part of the vocabulary corresponds to the question (or a paraphrase).

Hypotheses (2) and (4) are also supported by the empirical results. In cross-lingual experiments, the fact that the attention layers, connecting the context in one language to the question in another language, have allowed a focus on the location of the answer suggests that representations from both languages are similar. In addition, the fact that the results on Jap-Jap are better than on Jap-En suggests that the En-En SQuAD task learned by mBERT has been transferred to Jap-Jap by other mechanisms than language alignment alone, possibly by the inherent characteristics of the question-answering task that are language-invariant.

\section{KATE, a multilingual chatbot that answers from webpages}

The empirical results being very convincing, we integrated mBERT into our HR virtual assistant, KATE (Knowledge shAring experT for Employees). In this section, we present the use-case, the conversation process and how mBERT is integrated within the architecture. We also display some concrete examples of question and answers.

\begin{figure}[h]
\centering
\includegraphics[width=0.8\textwidth]{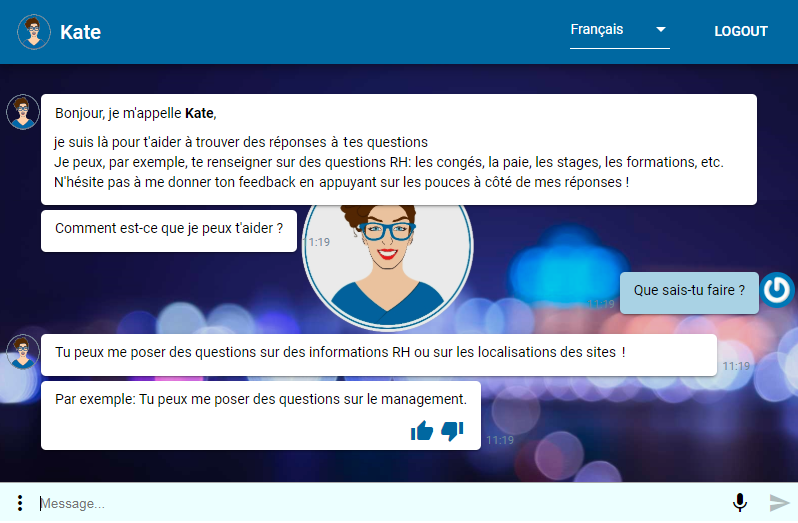}
\caption{Kate's Interface}\label{p4_interface_kate}
\end{figure}

Our application is a conversational assistant that answers human resources related questions. This has been motivated by an internal survey which highlighted how much employees struggle to find information in sources like an intranet, sometimes poorly structured and insufficiently indexed. Respondents indicated that these difficulties often lead to avoidance or abandonment. Therefore, we decided to create a chatbot with a web interface allowing standard text messaging interactions, and connected to a conversation framework developed in-house (figure \ref{p4_interface_kate}).

\begin{figure}[h]
\centering
\includegraphics[width=0.9\textwidth]{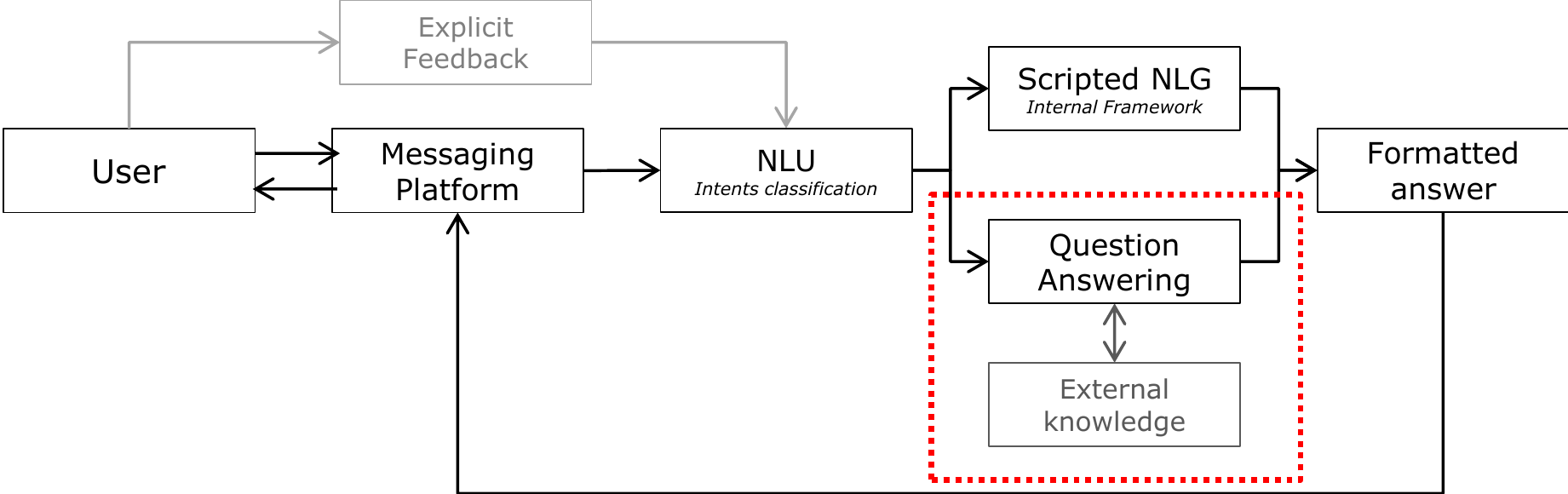}
\caption{Kate's conversational process (mBERT is integrated within the square in red dotted line)}\label{p4_fig_qa_KATE}
\end{figure}

Within Kate's conversational system (figure \ref{p4_fig_qa_KATE}), the messaging platform first sends the last user utterance to an intent classifier to handle basic questions (e.g. holidays policy, company sites location). For our experiment, we used Google Dialogflow NLU, but our framework could also handle other tools for this task like Rasa, Snips, Luis.ai or even BERT trained for classification. The dialog management is conducted within the framework. If the NLU engine succeeds to recognize the user's intent, we generate a scripted answer – basically, the developers prepared a set of pre-written answers to match the detected intent and the system chooses one of these sentences randomly and eventually completes it with the appropriate entities. The Question Answering part is called when the NLU engine fails, probably because the user asked for something that was not expected, or when we receive a negative feedback (see thumbs up/down buttons in figure \ref{p4_interface_kate}). In that case, we rely on our knowledge database, consisting of a list of URLs corresponding to the external contents that might include the right information – for our case, several corporate intranet webpages with HR explanations and rules. We consider the html code of each webpage, and apply a common HTML-to-text function to filter the HTML tags and keep only the actual content. Then we send these simplified texts to an instance of mBERT running on a server to perform the question answering, using the user utterance as a question and each text as the paragraph. In the last step, the best result among all sources is formatted and displayed by the messaging platform.

\begin{figure}[h]
\centering
\includegraphics[width=\textwidth]{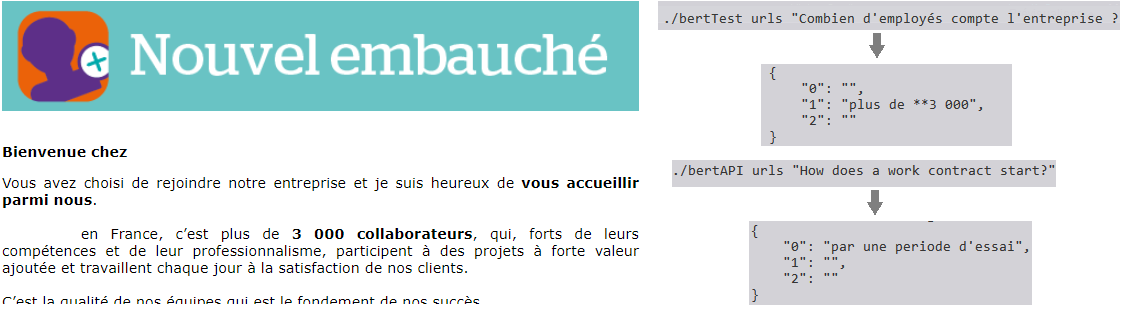}
\caption{Examples of responses of our QA API (right part of the figure). Here, the information sources used as context by BERT are three webs pages, one of which is shown on the left part.}\label{p4_example_qa_KATE_bert}
\end{figure}

In figure \ref{p4_example_qa_KATE_bert}, we give two examples of responses of our QA API, which runs with mBERT, using three urls as information sources. The first question, asked in french, is "\textit{How many employee does the company count?}" and the answer "\textit{more than 3000}" is found in the second web page. Note that the question and the context passage "\textit{en France, c'est plus de 3000 collaborateurs}" have no words in common. The second question "How does a work contract start ?" is asked in English whereas the source is in French. mBERT is still able to find the answer "par une periode d'essai" which translates into "with a trial period".

\section{Conclusion}

The results of mBERT on multilingual question answering tasks are impressive and confirm its aptitude for solving complex tasks and allowing zero-shot transfer. The provided answers seem very accurate, even when the question and the source have no word in common. This ability can be exploited to provide chatbots, in many languages, with the capacity to handle unexpected user questions using external information sources. This has huge maintenance benefits. Developers, linguists and chatbot integrators no longer need to manually define intentions/corpora/responses and can focus on updating the information sources and the language model. mBERT's cross-lingual skills provide an additional advantage for international companies who index documents in different languages.

On practical and generalization aspects, there is still room for improvement. First, mBERT is only able to return a partial answer when the information is dispersed in several parts of the source (subtitles, bulleted lists), so we are looking for ways to aggregate the several candidate responses. Second, the linguistic alignment of mBERT is limited for rare languages \cite{Pires2019HowMI} so we could benefit from an additional fine-tuning of the model on a parallel corpus (aligned pairs of sentences in English and the target languages). Finally, to scale up to very large sets of sources, we can investigate mBERT alternatives (e.g. distilbert, albert) in attempts to improve inference speed and memory consumption.
%
%
 \bibliographystyle{apalike}
 \bibliography{author.AKDM} 

\begin{thebibliography}{}

\bibitem[Agirre et~al., 2012]{agirre2012semeval}
Agirre, E., Diab, M., Cer, D., and Gonzalez-Agirre, A. (2012).
\newblock Semeval-2012 task 6: A pilot on semantic textual similarity.
\newblock In {\em Proceedings of the First Joint Conference on Lexical and
  Computational Semantics-Volume 1: Proceedings of the main conference and the
  shared task, and Volume 2: Proceedings of the Sixth International Workshop on
  Semantic Evaluation}, pages 385--393. Association for Computational
  Linguistics.

\bibitem[Asai et~al., 2018]{asai2018multilingual}
Asai, A., Eriguchi, A., Hashimoto, K., and Tsuruoka, Y. (2018).
\newblock Multilingual extractive reading comprehension by runtime machine
  translation.
\newblock {\em arXiv preprint arXiv:1809.03275}.

\bibitem[Devlin et~al., 2018]{devlin2018bert}
Devlin, J., Chang, M.-W., Lee, K., and Toutanova, K. (2018).
\newblock Bert: Pre-training of deep bidirectional transformers for language
  understanding.
\newblock {\em arXiv preprint arXiv:1810.04805}.

\bibitem[Dolan and Brockett, 2005]{dolan2005automatically}
Dolan, W.~B. and Brockett, C. (2005).
\newblock Automatically constructing a corpus of sentential paraphrases.
\newblock In {\em Proceedings of the Third International Workshop on
  Paraphrasing (IWP2005)}.

\bibitem[Firat et~al., 2016]{firat2016zero}
Firat, O., Sankaran, B., Al-Onaizan, Y., Vural, F. T.~Y., and Cho, K. (2016).
\newblock Zero-resource translation with multi-lingual neural machine
  translation.
\newblock {\em arXiv preprint arXiv:1606.04164}.

\bibitem[Hardalov et~al., 2019]{hardalov2019beyond}
Hardalov, M., Koychev, I., and Nakov, P. (2019).
\newblock Beyond english-only reading comprehension: Experiments in zero-shot
  multilingual transfer for bulgarian.
\newblock {\em arXiv preprint arXiv:1908.01519}.

\bibitem[Howard and Ruder, 2018]{howard2018universal}
Howard, J. and Ruder, S. (2018).
\newblock Universal language model fine-tuning for text classification.
\newblock {\em arXiv preprint arXiv:1801.06146}.

\bibitem[Johnson et~al., 2017]{johnson2017google}
Johnson, M., Schuster, M., Le, Q.~V., Krikun, M., Wu, Y., Chen, Z., Thorat, N.,
  Vi{\'e}gas, F., Wattenberg, M., Corrado, G., et~al. (2017).
\newblock Google’s multilingual neural machine translation system: Enabling
  zero-shot translation.
\newblock {\em Transactions of the Association for Computational Linguistics},
  5:339--351.

\bibitem[Jones, 2004]{jones2004statistical}
Jones, K.~S. (2004).
\newblock A statistical interpretation of term specificity and its application
  in retrieval.
\newblock {\em Journal of documentation}.

\bibitem[Joshi et~al., 2017]{joshi2017triviaqa}
Joshi, M., Choi, E., Weld, D.~S., and Zettlemoyer, L. (2017).
\newblock Triviaqa: A large scale distantly supervised challenge dataset for
  reading comprehension.
\newblock {\em arXiv preprint arXiv:1705.03551}.

\bibitem[Joulin et~al., 2016]{joulin2016bag}
Joulin, A., Grave, E., Bojanowski, P., and Mikolov, T. (2016).
\newblock Bag of tricks for efficient text classification.
\newblock {\em arXiv preprint arXiv:1607.01759}.

\bibitem[Kiros et~al., 2015]{kiros2015skip}
Kiros, R., Zhu, Y., Salakhutdinov, R.~R., Zemel, R., Urtasun, R., Torralba, A.,
  and Fidler, S. (2015).
\newblock Skip-thought vectors.
\newblock In {\em Advances in neural information processing systems}, pages
  3294--3302.

\bibitem[Lai et~al., 2017]{lai2017race}
Lai, G., Xie, Q., Liu, H., Yang, Y., and Hovy, E. (2017).
\newblock Race: Large-scale reading comprehension dataset from examinations.
\newblock {\em arXiv preprint arXiv:1704.04683}.

\bibitem[Le and Mikolov, 2014]{le2014distributed}
Le, Q. and Mikolov, T. (2014).
\newblock Distributed representations of sentences and documents.
\newblock In {\em International conference on machine learning}, pages
  1188--1196.

\bibitem[Levesque et~al., 2012]{levesque2012winograd}
Levesque, H., Davis, E., and Morgenstern, L. (2012).
\newblock The winograd schema challenge.
\newblock In {\em Thirteenth International Conference on the Principles of
  Knowledge Representation and Reasoning}.

\bibitem[Liu et~al., 2019a]{liu2019xqa}
Liu, J., Lin, Y., Liu, Z., and Sun, M. (2019a).
\newblock Xqa: A cross-lingual open-domain question answering dataset.
\newblock In {\em Proceedings of the 57th Conference of the Association for
  Computational Linguistics}, pages 2358--2368.

\bibitem[Liu et~al., 2019b]{liu2019roberta}
Liu, Y., Ott, M., Goyal, N., Du, J., Joshi, M., Chen, D., Levy, O., Lewis, M.,
  Zettlemoyer, L., and Stoyanov, V. (2019b).
\newblock Roberta: A robustly optimized bert pretraining approach.
\newblock {\em arXiv preprint arXiv:1907.11692}.

\bibitem[Logeswaran and Lee, 2018]{logeswaran2018efficient}
Logeswaran, L. and Lee, H. (2018).
\newblock An efficient framework for learning sentence representations.
\newblock {\em arXiv preprint arXiv:1803.02893}.

\bibitem[Manevitz and Yousef, 2001]{manevitz2001one}
Manevitz, L.~M. and Yousef, M. (2001).
\newblock One-class svms for document classification.
\newblock {\em Journal of machine Learning research}, 2(Dec):139--154.

\bibitem[Martin~d'Hoffschmidt, 2020]{fquad}
Martin~d'Hoffschmidt, Wacim~Belblidia, T. B. Q. H. M.~V. (2020).
\newblock Fquad: French question answering dataset.
\newblock {\em arXiv preprint arXiv:2002.06071}.

\bibitem[Mikolov et~al., 2013]{mikolov2013distributed}
Mikolov, T., Sutskever, I., Chen, K., Corrado, G.~S., and Dean, J. (2013).
\newblock Distributed representations of words and phrases and their
  compositionality.
\newblock In {\em Advances in neural information processing systems}, pages
  3111--3119.

\bibitem[Nguyen et~al., 2016]{nguyen2016ms}
Nguyen, T., Rosenberg, M., Song, X., Gao, J., Tiwary, S., Majumder, R., and
  Deng, L. (2016).
\newblock Ms marco: A human-generated machine reading comprehension dataset.

\bibitem[Pantel et~al., 1998]{pantel1998spamcop}
Pantel, P., Lin, D., et~al. (1998).
\newblock Spamcop: A spam classification \& organization program.
\newblock In {\em Proceedings of AAAI-98 Workshop on Learning for Text
  Categorization}, pages 95--98.

\bibitem[Peters et~al., 2018]{peters2018deep}
Peters, M.~E., Neumann, M., Iyyer, M., Gardner, M., Clark, C., Lee, K., and
  Zettlemoyer, L. (2018).
\newblock Deep contextualized word representations.
\newblock {\em arXiv preprint arXiv:1802.05365}.

\bibitem[Pires et~al., 2019]{Pires2019HowMI}
Pires, T., Schlinger, E., and Garrette, D. (2019).
\newblock How multilingual is multilingual bert?
\newblock In {\em ACL}.

\bibitem[Rajpurkar et~al., 2016]{rajpurkar2016squad}
Rajpurkar, P., Zhang, J., Lopyrev, K., and Liang, P. (2016).
\newblock Squad: 100,000+ questions for machine comprehension of text.
\newblock {\em arXiv preprint arXiv:1606.05250}.

\bibitem[Schuster and Nakajima, 2012]{schuster2012japanese}
Schuster, M. and Nakajima, K. (2012).
\newblock Japanese and korean voice search.
\newblock In {\em 2012 IEEE International Conference on Acoustics, Speech and
  Signal Processing (ICASSP)}, pages 5149--5152. IEEE.

\bibitem[Socher et~al., 2013]{socher2013recursive}
Socher, R., Perelygin, A., Wu, J., Chuang, J., Manning, C.~D., Ng, A., and
  Potts, C. (2013).
\newblock Recursive deep models for semantic compositionality over a sentiment
  treebank.
\newblock In {\em Proceedings of the 2013 conference on empirical methods in
  natural language processing}, pages 1631--1642.

\bibitem[Vaswani et~al., 2017]{vaswani2017attention}
Vaswani, A., Shazeer, N., Parmar, N., Uszkoreit, J., Jones, L., Gomez, A.~N.,
  Kaiser, {\L}., and Polosukhin, I. (2017).
\newblock Attention is all you need.
\newblock In {\em Advances in neural information processing systems}, pages
  5998--6008.

\bibitem[Wang et~al., 2018]{wang2018glue}
Wang, A., Singh, A., Michael, J., Hill, F., Levy, O., and Bowman, S.~R. (2018).
\newblock Glue: A multi-task benchmark and analysis platform for natural
  language understanding.
\newblock {\em arXiv preprint arXiv:1804.07461}.

\bibitem[Warstadt et~al., 2018]{warstadt2018neural}
Warstadt, A., Singh, A., and Bowman, S.~R. (2018).
\newblock Neural network acceptability judgments.
\newblock {\em arXiv preprint arXiv:1805.12471}.

\bibitem[Williams et~al., 2017]{williams2017broad}
Williams, A., Nangia, N., and Bowman, S.~R. (2017).
\newblock A broad-coverage challenge corpus for sentence understanding through
  inference.
\newblock {\em arXiv preprint arXiv:1704.05426}.

\bibitem[Yang et~al., 2019]{yang2019xlnet}
Yang, Z., Dai, Z., Yang, Y., Carbonell, J., Salakhutdinov, R., and Le, Q.~V.
  (2019).
\newblock Xlnet: Generalized autoregressive pretraining for language
  understanding.
\newblock {\em arXiv preprint arXiv:1906.08237}.

\end{thebibliography}
%


\end{document}